\documentclass[11pt]{article}

\usepackage{emnlp2021}

\usepackage{times}
\usepackage{latexsym}
\usepackage{booktabs}

\usepackage[T1]{fontenc}

\usepackage[utf8]{inputenc}

\usepackage{microtype}

\newcommand{\Picard}{\textsc{Picard}}
\newcommand{\texttosql}{text-to-SQL}

\usepackage{tikz}
\def\checkmark{\tikz\fill[scale=0.35](0,.35) -- (.2,0) -- (.65,.7) -- (.2,.15) -- cycle;}

\title{%
\Picard{}:
\\ Parsing Incrementally for Constrained Auto-Regressive Decoding
\\ from Language Models%
}

\author{%
Torsten Scholak \and
Nathan Schucher \and
Dzmitry Bahdanau \\
ElementAI, a ServiceNow company \\
\texttt{\{torsten.scholak,dzmitry.bahdanau\}@servicenow.com}%
}

\begin{document}
\maketitle
\begin{abstract}
Large pre-trained language models for textual data have
an unconstrained output space;
at each decoding step,
they can produce any of 10,000s of sub-word tokens.
When fine-tuned to target constrained formal languages like SQL,
these models often generate invalid code, rendering it unusable.
We propose \Picard{}\footnote{The \Picard{} code is available at
\url{https://github.com/ElementAI/picard}.},
a method for constraining auto-regressive decoders of language models
through incremental parsing.
\Picard{} helps to find valid output sequences by rejecting inadmissible tokens
at each decoding step.
On the challenging Spider and CoSQL \texttosql{} translation tasks,
we show that \Picard{} transforms fine-tuned T5 models
with passable performance into state-of-the-art solutions.
\end{abstract}

\section{Introduction}

While there have been many successes in applying large pre-trained language models
to downstream tasks,
our ability to control and constrain the output of these models is still very limited.
Many enterprise applications are out of reach
because they require a degree of rigour and exactitude that
language models are not able to deliver yet.
If the target is a formal language like SQL,
then we would like the model to adhere exactly and provably to the SQL specification
with all its lexical, grammatical, logical, and semantical constraints.
Unfortunately, with pre-training alone, language models may not satisfy
these correctness requirements.

For \texttosql{} translation,
the most widespread solution to constrained decoding
is to make invalid SQL unrepresentable.
For a while now
it has been possible to restrict auto-regressive decoding to only those token sequences
that correctly parse to SQL abstract syntax trees
\citep{Yin_2018,lin2019grammarbased,Wang_2020}.
More recently,
semi-auto-regressive improvements to this parsing paradigm
have been proposed \citep{rubin2020smbop}.
However, while effective,
these approaches have in common that they are achieved at the expense of
using a custom vocabulary of special control tokens
or a custom model architecture, or both.
Unfortunately,
this makes them incompatible with generic pre-trained language model decoders.
A less invasive and more compatible approach is
to not constrain the generation process,
but instead to filter finalized beam hypotheses by validity
\citep{suhr-etal-2020-exploring,Lin_2020}.
Yet, such filtering is at the expense of a very large beam size.

\begin{figure}
\centering
\small
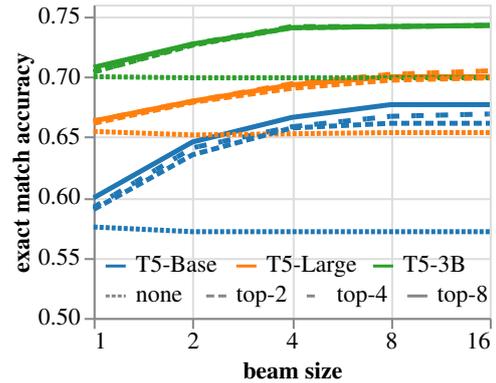
\caption{Exact-set-match accuracy of the highest-scoring prediction
as a function of beam size on the Spider \texttosql{} development set.
With \Picard{} turned on,
token predictions had to pass \Picard{} checking
at every decoding step.
Only the top-2, -4, and -8 token predictions
of each hypothesis were considered in the beam search.
With \Picard{} turned off (none),
all token predictions were considered and none were checked.
The models, T5-Base, -Large, and -3B, did not have access to any database content,
only to the database schemas.}
\label{fig:beam_size}
\end{figure}

We address the expenses of these approaches with
a novel incremental parsing method
for constrained decoding called \Picard{},
which stands for
"Parsing Incrementally for Constrained Auto-Regressive Decoding."
\Picard{} is compatible with any existing auto-regressive language model decoder
and vocabulary---%
including, but not limited to, those of large pre-trained transformers---%
and it does not require very large beam sizes.
\Picard{} is entirely absent from pre-training or fine-tuning of the model,
and can be easily and optionally enabled at inference time.
\Picard{} operates directly on the output of the language model
which, in the case of \texttosql{} translation,
is the readable surface form of the SQL code.

In our experiments, we find that
\Picard{} can significantly improve the performance of a large pre-trained language model
\citep{t5}
after it is fine-tuned on the \texttosql{} task.
On the Spider \texttosql{} dataset \citep{Yu_2018},
we find that a T5-Base model with \Picard{} can outperform a T5-Large model without it,
and likewise for a T5-Large and a T5-3B model.
Significantly,
with the help of \Picard{},
a T5-3B model can be raised to state-of-the-art performance on
the Spider and CoSQL datasets \citep{yu-etal-2019-cosql}.

\section{The \Picard{} Method}

\begin{figure}
\centering
\small
\def\svgwidth{.8\columnwidth}
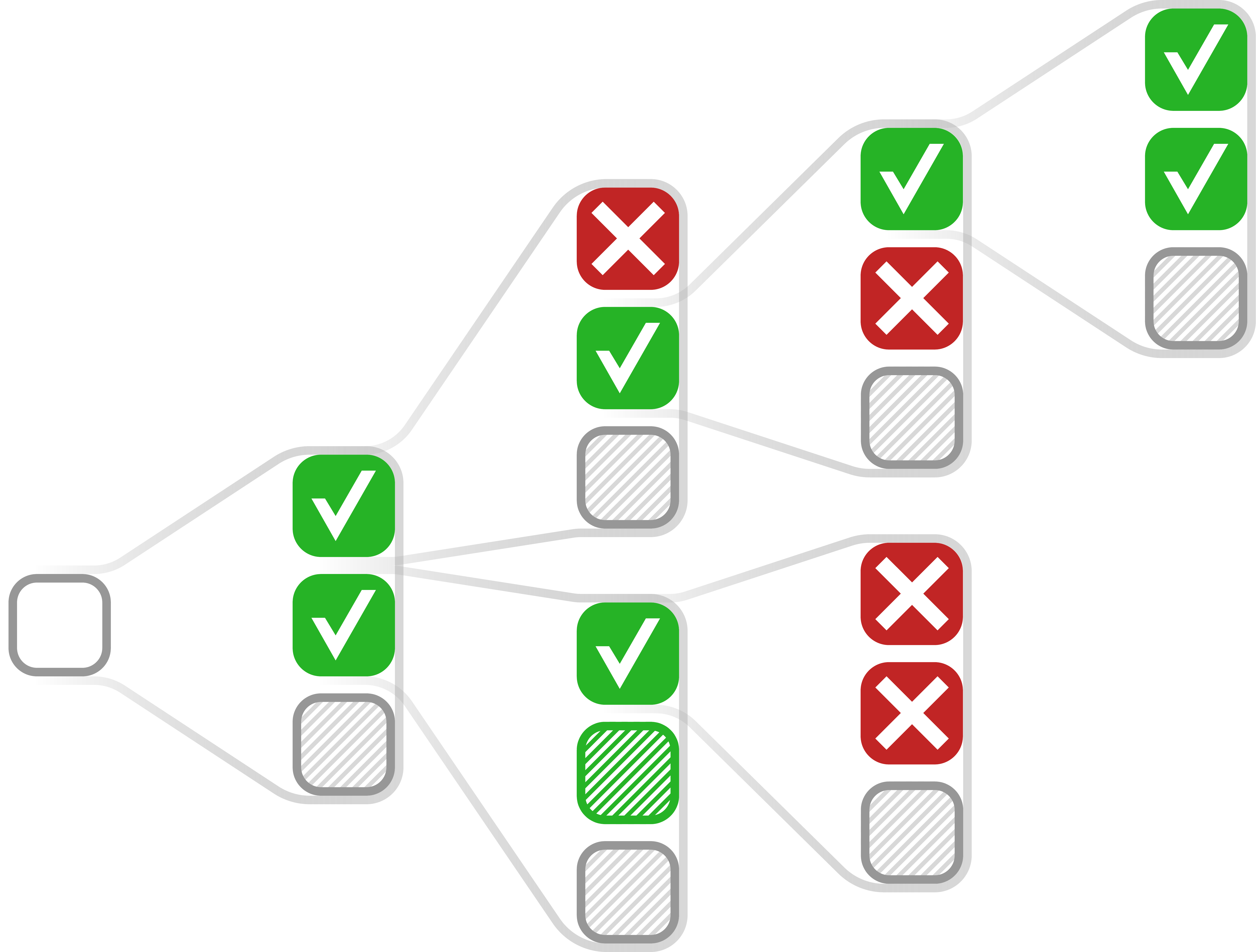
\caption{%
Illustration of constrained beam search with beam size $2$
and \Picard{}.
Each vertical column represents three token predictions for a hypothesis
from top to bottom in descending order by probability.
In this example, \Picard{} is configured to only check
the top-2 highest ones.
The rest is automatically dismissed by setting their score
to $-\infty$.
Tokens rejected by \Picard{} (red, $\times$)
are also assigned a score of $-\infty$.
Accepted tokens (green, \checkmark{}) keep their original score.%
}
\label{fig:picard_explanation}
\end{figure}

\Picard{} warps model prediction scores
and integrates trivially with existing algorithms for
greedy and beam search used in
auto-regressive decoding from language models.
Its arguments are the token ids of the current hypothesis
and, for each vocabulary token, the log-softmax scores
predicted by the model's language modeling head.
\Picard{} also has access to SQL schema information,
in particular, information about the names of tables and columns
and about which column resides in which table.

At each generation step,
\Picard{} first restricts prediction to
the top-$k$ highest probability tokens
and then assigns a score of $-\infty$ to those
that fail \Picard{}'s numerous checks
(see Figure~\ref{fig:picard_explanation}).
These checks are enabled by fast incremental parsing \cite{attoparsec}
based on monadic combinators \cite{leijen2001parsec}.
There are four \Picard{} mode settings that
control their comprehensiveness:
off (no checking),
lexing,
parsing without guards, and
parsing with guards---the highest mode.
A prediction that passes a higher mode will always pass a lower mode
but not necessarily vice versa.

\subsection{Lexing}

In lexing mode, \Picard{} checks the output on a lexical level only.
It attempts to convert the partial, detokenized model output to
a white-space delimited sequence of individual
SQL keywords like \verb+select+,
punctuation like \verb+()+,
operators like \verb-+- and \verb+-+,
literals like string and number values in SQL conditions,
and identifiers like aliases, tables, and columns---%
without being sensitive to the order in which these lexical items appear.
By making it so, \Picard{} can detect spelling errors in keywords
or reject table and column names that are invalid for the given SQL schema.
For instance, consider the question
"What are the email, cell phone and home phone of each professional?"
from Spider's development set on the \verb+dog_kennels+ database.
Our fine-tuned T5-Large model predicts
\verb+select email_address, cell_phone,+
\verb+home_phone from professionals+
while the ground truth selects \verb+cell_number+ instead of
the invalid \verb+cell_phone+ column.
This mistake is caught and avoided by \Picard{} in lexing mode.

\subsection{Parsing without Guards}

In the lowest parsing mode above lexing---referred to as parsing without guards---%
\Picard{} checks the output on a grammatical level.
\Picard{} attempts to parse the detokenized model output to a data structure
that represents the abstract syntax tree (AST) of the predicted SQL query.
Contrary to lexing mode,
the order in which keywords and clauses appear now matters.
\Picard{} can reject invalid query structures,
e.g.\ find missing \verb+from+ clauses or incorrect orders of clauses and keywords.
It can also detect a range of issues with compositions of SQL expressions:
Number one,
if \Picard{} matches on a \verb+tid.cid+ pattern,
but the table with the id \verb+tid+ does not contain
a column with id \verb+cid+, then that parse is rejected.
Secondly,
if \Picard{} first matches on an \verb+alias.cid+ pattern
and then later matches on the \verb+tid as alias+ pattern
but \verb+tid+ does not contain \verb+cid+, then that parse is also rejected.
An equivalent rule also exists for sub-queries bound to table aliases.
Lastly,
\Picard{} prohibits duplicate binding of a table alias
in the same \verb+select+ scope, but permits shadowing of
aliases defined in a surrounding scope.
This can happen in nested SQL queries.

\subsection{Parsing with Guards}

In its highest parsing mode,
\Picard{} engages in additional analyses---called guards---%
while assembling the SQL AST.
If \Picard{} matches on \verb+tid.cid+ or \verb+alias.cid+,
then guards require that the table \verb+tid+ or the alias \verb+alias+, respectively,
is eventually brought into scope
by adding it to the \verb+from+ clause.
Moreover,
the alias \verb+alias+ is constrained to resolve to a table or a sub-query
that has the column \verb+cid+ in it.
If \Picard{} matches on the pattern \verb+cid+,
then another guard requires that
exactly one table is eventually brought into scope
that contains a column with that id.
These guards are enforced eagerly
in order to fail fast and to eject invalid hypotheses
from the beam at the earliest possible time.
The first time this is happening is after parsing the \verb+from+ clause.

Only with these guards,
\Picard{} is able to reject a wrong prediction from our fine-tuned T5-Large model like 
\verb+select maker, model+ \verb+from car_makers+
for the question "What are the makers and models?"
Here, the correct table to use would have been \verb+model_list+,
since it is the only one in Spider's \verb+car_1+ schema that
contains both a \verb+maker+ and a \verb+model+ column.

Additional checks and guards are conceivable,
for instance, checking that only expressions of the same type are compared
or that column types selected by \verb+union+, \verb+except+, or \verb+intersect+
queries match.
We leave these additional checks to future work.

\section{Experiments}

Our experiments are mainly
focused on Spider \citep{Yu_2018},
a large multi-domain and cross-database dataset for \texttosql{} parsing.
We train on the 7,000 examples in the Spider training set
and evaluate on Spider's development set and its hidden test set.
We also report results on the
CoSQL SQL-grounded dialog state tracking task \citep{yu-etal-2019-cosql},
where we predict a SQL query for each question given previous questions
in an interaction context.
For this task, we train on both the Spider \texttosql{} training data
and the CoSQL dialog state tracking training data,
and evaluate on the CoSQL development and test sets.

Spider and CoSQL are both zero-shot settings.
There is no overlap between questions or databases
between the respective training, development, and test sets.

On Spider, we determine model performance based on three metrics:
exact-set-match accuracy, execution accuracy,
and test-suite execution accuracy \citep{Zhong_2020}.
Exact-set-match accuracy
compares the predicted and the ground-truth SQL query
by parsing both into a normalized data structure.
This comparison is not sensitive to literal query values
and can decrease under semantic-preserving SQL query rewriting.
Execution accuracy compares the results of
executing the predicted and ground-truth SQL queries on
the database contents shipped with the Spider dataset.
This metric is sensitive to literal query values,
but suffers from a high false positive rate \citep{Zhong_2020}.
Lastly, test-suite execution accuracy 
extends execution to multiple database instances per SQL schema.
The contents of these instances
are optimized to lower the number of false positives
and to provide the best approximation of semantic accuracy.

On CoSQL, we measure model performance in terms of
the question match accuracy and the interaction match accuracy.
Both metrics are based on exact-set-match accuracy.
Interaction match accuracy is the joint accuracy over all questions in an interaction.

We are encouraged by results by \citet{shaw2020compositional},
who showed that a pre-trained T5-Base or T5-3B model
can not only learn the \texttosql{} task,
but also generalize to unseen databases,
and even that T5-3B can be competitive
with the then-state-of-the-art \citep{Choi_2021,Wang_2020}---%
all without modifications to the model.
We therefore use T5 as the baseline for all our experiments.

In order to allow for
generalization to unseen databases,
we encode the schema together with the questions.
We use the same serialization scheme used by \citet{shaw2020compositional}.
In experiments using database content,
we detect and attach the database values to the column names
in a fashion similar to the BRIDGE model by \citet{Lin_2020}.
When fine-tuning for the CoSQL dialog state tracking task,
we append the previous questions in the interaction in reverse chronological order to the input.
Inputs exceeding the 512-token limit of T5 are truncated.
The target is the SQL from the Spider and/or CoSQL training sets,
unmodified except for a conversion of keywords and identifiers to lower case.
We fine-tune T5 for up to $3072$ epochs
using Adafactor \citep{shazeer2018adafactor},
a batch size of $2048$,
and a learning rate of $10^{-4}$.

\begin{table*}[ht]
\centering
\small
\setlength{\tabcolsep}{6pt}
\begin{tabular}{lcccc}
\toprule
                                                                      & \multicolumn{2}{c}{Development} & \multicolumn{2}{c}{Test}        \\
\cmidrule{2-5}
System                                                                & EM\%           & EX\%           & EM\%           & EX\%           \\
\midrule
BRIDGE v2 + BERT (ensemble)$^\dagger$ \citep{Lin_2020}                & 71.1           & 70.3           & 67.5           & 68.3           \\
\textsc{SmBoP} + \textsc{GraPPa}$^\dagger$ \citep{rubin2020smbop}     & 74.7           & 75.0           & 69.5           & 71.1           \\
RATSQL + \textsc{GAP}$^\dagger$ \citep{shi2021learning}               & 71.8           & -              & 69.7           & -              \\
DT-Fixup SQL-SP + \textsc{RoBERTa}$^\dagger$ \citep{xu2020optimizing} & 75.0           & -              & 70.9           & -              \\
LGESQL + \textsc{ELECTRA}$^\dagger$ \citep{cao-etal-2021-lgesql}      & 75.1           & -              & \textbf{72.0}  & -              \\
T5-Base \cite{shaw2020compositional}                                  & 57.1           & -              & -              & -              \\
T5-3B \cite{shaw2020compositional}                                    & 70.0           & -              & -              & -              \\
\midrule
T5-Base (ours)                                                        & 57.2           & 57.9           & -              & -              \\
T5-Base+\Picard{}                                                     & 65.8           & 68.4           & -              & -              \\
T5-Large                                                              & 65.3           & 67.2           & -              & -              \\
T5-Large+\Picard{}                                                    & 69.1           & 72.9           & -              & -              \\
T5-3B (ours)                                                          & 69.9           & 71.4           & -              & -              \\
T5-3B+\Picard{}                                                       & 74.1           & 76.3           & -              & -              \\
T5-3B$^\dagger$                                                       & 71.5           & 74.4           & 68.0           & 70.1           \\
T5-3B+\Picard{}$^\dagger$                                             & \textbf{75.5}  & \textbf{79.3}  & \textbf{71.9}  & \textbf{75.1}  \\
\bottomrule
\end{tabular}
\caption{Our results (bottom) and relevant prior art (top) on the Spider \texttosql{} task. Shown are the exact-set-match accuracy (EM) and execution accuracy (EX) percentages on Spider's development and test sets. Our results are for a beam of size 4, and \Picard{} is parsing with guards for the top-2 token predictions. A dagger (\textdagger{}) indicates use of database content, otherwise schema only.}
\label{tbl:main_results}
\end{table*}

\paragraph{Results}

Our findings on the Spider dataset are summarized in Table~\ref{tbl:main_results}
and Figure~\ref{fig:beam_size}.
Our reproductions of \citet{shaw2020compositional}'s results
with T5 cannot compete with the current state of the art on Spider.
The issue is that these models predict a lot of invalid SQL.
For instance, 12\% of the SQL queries generated by the T5-3B model
on Spider's development set result in an execution error.
However, when these same models are augmented with \Picard{},
we find substantial improvements.
First, invalid SQL predictions become rare.
For T5-3B with \Picard{},
only 2\% of the predictions are unusable.
In these cases,
beam search exited without finding a valid SQL prediction.
Second, and most significantly,
by using \Picard{}, the T5-3B model is lifted to state-of-the-art performance.
We measure an exact-set-match accuracy of
75.5\% on the development set and 71.9\% on the test set.
The execution accuracy results are 79.3\% and 75.1\%, respectively.
These numbers are on par or higher than those of the closest competitor,
LGESQL + \textsc{ELECTRA} \citep{cao-etal-2021-lgesql} (see Table~\ref{tbl:main_results}).
Furthermore, we achieve a test-suite execution accuracy of 71.9\% on Spider's development set.

Our findings on the CoSQL dialog state tracking dataset
(see Table~\ref{tbl:cosql_results})
are similar to those for Spider.
\Picard{} significantly improves the performance,
and our fine-tuned T5-3B model achieves state-of-the-art performance.

\Picard{} is not only improving performance, it is also fast.
During evaluation of the T5-3B model on Spider,
the decoding speed with beam size $4$ on an NVIDIA A100-SXM4-40GB GPU
was, on average, 2.5 seconds per sample without \Picard{}
and 3.1 seconds per sample with \Picard{}.

\begin{table*}
\centering
\small
\setlength{\tabcolsep}{6pt}
\begin{tabular}{lcccc}
\toprule
                                                                      & \multicolumn{2}{c}{Development} & \multicolumn{2}{c}{Test}        \\
\cmidrule{2-5}
System                                                                & QM\%           & IM\%           & QM\%           & IM\%           \\
\midrule
RATSQL + \textsc{SCoRe} \citep{yu2021score}                           & 52.1           & 22.0           & 51.6           & 21.2           \\
\midrule
T5-3B                                                                 & 53.8           & 21.8           & 51.4           & 21.7           \\
T5-3B+\Picard{}                                                       & \textbf{56.9}  & \textbf{24.2}  & \textbf{54.6}  & \textbf{23.7}  \\
\bottomrule
\end{tabular}
\caption{Our results (bottom) and relevant prior art (top) on the CoSQL dialog state tracking task. Shown are the question match accuracy (QM) and interaction match accuracy (IM) percentages on CoSQL's development and test sets. Our results are for a beam of size 4, and \Picard{} is parsing with guards for the top-2 token predictions.}
\label{tbl:cosql_results}
\end{table*}

\paragraph{Beam Size}

Figure~\ref{fig:beam_size}
shows results on Spider without and with \Picard{} when parsing with guards
for different beam sizes and sizes of T5.
For each model size,
\Picard{} increases performance with increasing beam size.
These increases are the strongest for the step from beam size $1$ to $2$,
less pronounced from $2$ to $4$,
and then saturating for beam sizes above $4$.
Even with greedy search (beam size $1$),
\Picard{} allows for some modest improvements.
Note that, without \Picard{}, these models do not benefit from beam search.
The number, $k$, of highest-probability tokens that are processed by \Picard{}
at each decoding step has a modest to negligible impact on performance.
It is the largest for T5-Base,
smaller for T5-Large, and almost undetectable for T5-3B.
We do not study the case $k = 1$,
because it reduces the beam search to constrained greedy search.

\paragraph{Ablations}

\begin{figure}
\centering
\small
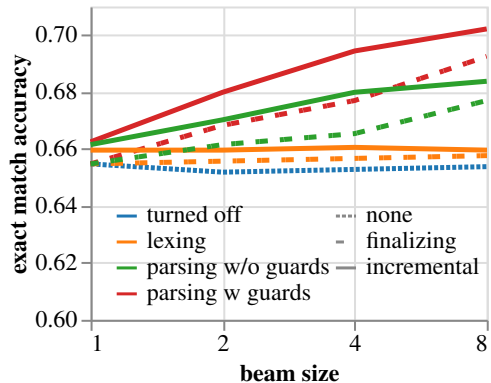
\caption{%
Exact-set-match accuracy on the Spider development set as a function of beam size
for top-$4$ \Picard{} on T5-Large (schema only) and for different operation modes:
turned off, lexing, parsing without guards, and parsing with guards.
In each mode, \Picard{} is either used incrementally at each step or
only when finalizing a hypothesis.%
}
\label{fig:picard_mode}
\end{figure}

In Figure~\ref{fig:picard_mode},
we have condensed our ablation analysis for \Picard{}.
We show results for our T5-Large model in all four \Picard{} checking modes
and for four different beam sizes on the Spider development set.
When checking incrementally at each decoding step,
lexing shows a small improvement over the unconstrained T5 model.
The results without \Picard{} and with \Picard{} in lexing mode
are largely independent of the beam size.
This is different when \Picard{} is switched into
the more sophisticated parsing modes.
Both, with and without guards, improvements from \Picard{} increase
rapidly for increasing beam sizes,
where parsing with guards clearly has a strong lead over parsing without them.

In order to compare \Picard{} with the filtering-by-validity approach of
\citet{suhr-etal-2020-exploring} and \citet{Lin_2020},
we have studied also what happens when \Picard{} is only checking
hypotheses when the model predicts their finalization with the end-of-sequence token.%
\footnote{This is not exactly equivalent to filtering a completely finalized beam,
because the hypotheses rejected by \Picard{} never enter it and never take up any space.}
In this restrained mode,
\Picard{} is still effective, but much less so compared to normal incremental operation.
The gap between these two modes of operation only begins to shrink
for large beam sizes.
This is understandable since \citet{Lin_2020} used beam sizes of at least $16$ and up to $64$
to reach optimal results with filtering while \citet{suhr-etal-2020-exploring}
used a beam of size $100$.

\section{Conclusion}

We propose and evaluate a new method, \Picard{},
for simple and effective constrained decoding with large pre-trained language models.
On both, the Spider cross-domain and cross-database \texttosql{} dataset
and the CoSQL SQL-grounded dialog state tracking dataset,
we find that the \Picard{} decoding method not only significantly improves
the performance of fine-tuned but otherwise unmodified T5 models,
it also lifts a T5-3B model to state-of-the-art results
on the established exact-match and execution accuracy metrics.

\section*{Acknowledgements}

We thank Lee Zamparo for his contributions to the experiments on the CoSQL dataset.
Further, we would like to thank Pete Shaw
for his input on the reproduction of the T5 results on Spider.
We would also like to extend our gratitude to Tao Yu and Yusen Zhang for their
efforts in evaluating our model on the test split of the Spider
and CoSQL datasets.
Finally, we thank our anonymous reviewers for their time and valuable suggestions.

\bibliography{custom}
\bibliographystyle{acl_natbib}

\end{document}